\documentclass{article}
\usepackage{arxiv,epsfig}
\usepackage{siunitx}
\usepackage{amsmath, amssymb, amsfonts, amsbsy}

\usepackage{subfig}

\usepackage{cite} %
\renewcommand{\baselinestretch}{0.97}

\newcommand{\x}{{\mathbf x}}

\usepackage{color}

\usepackage{hyperref}
\hypersetup{
    bookmarks=true,         %
    unicode=false,          %
    pdftoolbar=true,        %
    pdfmenubar=true,        %
    pdffitwindow=false,     %
    pdfstartview={FitH},    %
    pdftitle={Uncertainty},    %
    pdfauthor={Camps-Valls},     %
    pdfsubject={Uncertainty},   %
    pdfcreator={Camps-Valls},   %
    pdfproducer={Camps-Valls}, %
    pdfkeywords={keyword1} {key2} {key3}, %
    pdfnewwindow=true,      %
    colorlinks=true,       %
    linkcolor=blue,          %
    citecolor=blue,        %
    filecolor=blue,      %
    urlcolor=blue           %
}

\title{Disentangling Derivatives, Uncertainty and Error \\ in Gaussian Process Models}

\author{
  J. Emmanuel Johnson \\
  Image Processing Laboratory \\
  Universitat de Val{\`e}ncia\\
  Val{\`e}ncia, Spain\\
  \texttt{juan.johnson@uv.es} \\
  \And
  Valero Laparra \\
  Image Processing Laboratory \\
  Universitat de Val{\`e}ncia\\
  Val{\`e}ncia, Spain\\
  \texttt{valero.laparra@uv.es} \\
  \And
  Gustau Camps-Valls \\
  Image Processing Laboratory \\
  Universitat de Val{\`e}ncia\\
  Val{\`e}ncia, Spain\\
  \texttt{gcamps@uv.es} \\
}

\begin{document}

\begin{center}
    ©IEEE. ACCEPTED FOR PUBLICATION IN IEEE IGARSS 2018. DOI 10.1109/IGARSS.2018.8519020\footnote{
©IEEE. Personal use of this material is permitted.  Permission from IEEE must be obtained for all other users,including reprinting/republishing this material for advertising or promotional purposes, creating new collective works for resale or redistribution to servers or lists, or reuse of any copyrighted components of this work in other works.  DOI: 10.1109/IGARSS.2018.8519020.
}
\end{center}

\maketitle

\begin{abstract}
Gaussian Processes (GPs) are a class of kernel methods that have shown to be very useful in geoscience applications. They are widely used because they are simple, flexible and provide very accurate estimates for nonlinear problems, especially in parameter retrieval. An addition to a predictive mean function, GPs come equipped with a useful property: the predictive variance function which provides confidence intervals for the predictions. The GP formulation usually assumes that there is no input noise in the training and testing points, only in the observations. However, this is often not the case in Earth observation problems where an accurate assessment of the instrument error is usually available. In this paper, we showcase how the derivative of a GP model can be used to provide an analytical error propagation formulation and we analyze the predictive variance and the propagated error terms in a temperature prediction problem from infrared sounding data.
\end{abstract}


\section{Introduction}

The land and sea surface temperature of the Earth is one of the most important components to understanding the governing physical processes on the Earth's surface \cite{blondeau-patissier_review_2014}. Derived processes such as heat-fluxes and energy balances on a large temporal and spatial scale are useful for applications within climate change, vegetation monitoring and other environmental studies. In order to acquire a complete model for many of these applications, one needs a good characterization of temperature on a global scale \cite{li_satellite-derived_2013}. Acquiring ground measurements is not always practical at such a high spatial and temporal resolution scale. Remote sensing has proven to be useful to collect input data for models that capture temperature and other important environmental factors. Instruments such as the Infrared Atmospheric Sounding Interferometer (IASI)~\cite{EUMETSAT-IASI-MISSION}, have an objective to support numerical weather predictions (NWP) models to provide high quality predictions for temperature, humidity, and some trace gases. Generation of global maps from satellite acquisitions %
do require fast and accurate algorithms. %

It is standard practice to use physics-based models, statistical-based models or a mixture of both~\cite{CampsValls2011mc,campsvalls11eum}. In this paper, we will focus on a subclass of statistical-based models, Gaussian processes (GPs)~\cite{rasmussen_gaussian_2006}, which have been shown to be very useful in remote sensing applications \cite{camps-valls_survey_2016}. A particular advantage of GPs is that they provide the predictive variance of the modeled data, as it provides a good measure of the confidence in the predictions. Another property of GPs is that one can easily calculate the derivative of the predictions as well as the derivative of the predictive variance \cite{rasmussen_gaussian_2006}. The derivative is also related to the uncertainty estimation by providing access to error propagation estimates. Although it is apparent that both the predictive variance and the derivative of the kernel model can be used to acquire error estimates, the exact relationship between these features is still not fully understood or utilized within the machine learning or applied communities. The error estimates of any model is becoming increasingly more and more important as models get more involved and complex, and it is clear that the predictive variance alone does not take into account the noise in the inputs for a GP model. Thus if the machine learning community hopes to incorporate more statistical learning models (like the GP model) to supplement complex physical models, then one needs to have accurate error estimates. 

The rest of the paper is outlined as follows. \S2 fixes the notation and the main concepts of error and uncertainty propagation. We explore the relations between the measurements error, the model derivatives, and the predictive variance, and show how they relate to the final error of the model. \S3 showcases an example of application for temperature prediction from IASI data. We conclude in \S4 with some remarks and future work.

\section{Methods}

Here we give a basic overview of Gaussian process regression (GP) regression followed by the derivative of the GP model, how this can used to calculate the propagation of the input noise and included in the predictive variance function to incorporate errors of the inputs. 

\subsection{Gaussian Process Regression}

We are given $N$ pairs of input-output points, $\{\x_i,y_i|i=1,\ldots,N\}$, where $\x\in\mathbb{R}^{D\times 1}$ and $y\in\mathbb{R}$. Let us define $\mathbf{X} =\left[\x_1|\cdots| \x_n \right]^\intercal\in\mathbb{R}^{N \times D}$ be a set of known $N$ data points, and $\mathbf{y}=\left[y_1, y_2, \ldots, y_n \right]\in\mathbb{R}^{N\times 1}$ be the known $N$ labels in $\mathbb{R}^{N \times 1}$. We are interested in finding a latent function $f(\x)$ of input $\x$ that approximates $y$.

GPs are a popular kernel-based method for discriminative learning \cite{williams_gaussian_1996,rasmussen_gaussian_2006}. GPs use principles from the Bayesian regression framework where a function $f(\x)$ is defined to describe the variable of interest $y$ under an additive noise model
\begin{equation}\label{eq:signalmodel}
y = f(\x) + \epsilon_{y},
\end{equation}
where $\epsilon_y$ represents the modeling error or residuals. By assuming a Gaussian prior for the noise term $\epsilon_y\sim\mathcal{N}({\bf 0}, \sigma^2_y)$, and a zero mean GP prior for the latent function, $f(\x) \sim \mathcal{GP} \left( \mathbf{0}, \mathbf{K} \right)$, where $\mathbf{K}$ is the covariance matrix parameterized using a kernel function, $K(\x_i,\x_j)$, we can analytically compute the posterior distribution over the unknown output $y_*$, $p\left(y_*|x_*, \mathcal{D}\right) = \mathcal{N}(y_* | \mu_*, \sigma_*^2)$ with the following predictive mean and variance for a new incoming test input point $\x_*$:
\begin{equation}
	\mu_* = {\bf k}_* ({\bf K}+\lambda {\bf I}_N)^{-1}{\bf y} = {\bf k}_* \alpha 
        \label{eq:gp_pred}
\end{equation}
\begin{equation}
    \sigma_*^2 = \sigma_y^2 + k_{**}- {\bf k}_* ({\bf K}+\sigma_y^2 \mathbf{I}_N)^{-1} {\bf k}_{*}^{\top},
    \label{eq:gp_var}
\end{equation}
where %
${\bf k}_*=[K(\x_*,\x_1),\ldots,K(\x_*,\x_N)]^\intercal\in\mathbb{R}^{N\times 1}$ contains the similarities between the test point $\x_*$ and all the training samples, $k_**$ is the self-similarity for the test sample (a scalar). %
Note how computing the predictive variance, $\sigma_*$, is straightforward. The expression above is the same as Kernel Ridge Regression (KRR) or the Relevance Vector Machine (RVM) with the primary difference being how to find the kernel function hyper-parameters.  

\subsection{Input-Noise and Derivatives}

In GPs, normally we have two standard assumptions: (1) the outputs $y$ are corrupted by the constant-variance Gaussian noise; and (2) the input vectors $x$ are noise-free. These are assumptions of practical convenience, but far from being realistic in real problems. %
Actually, in remote sensing, we often have the case where the input measurements are corrupted with noise especially when the data is acquired using instruments, atmospherically and geometrically corrected and transformed, or derived from realistic models. 

Let $\tilde{\x}$ be the observed vector that contains the real input vector $\x$ corrupted by some AWGN $\boldsymbol{\epsilon}_x\sim{\mathcal N}({\bf 0},\boldsymbol{\Sigma}_x)$: %
\begin{equation}\label{eq:inputnoisemodel}
	\tilde{\x}=\x + \boldsymbol{\epsilon}_x.
\end{equation}
Then introducing \eqref{eq:inputnoisemodel} in \eqref{eq:signalmodel}, we simply obtain:
\begin{equation*}
	y=f(\x+\boldsymbol{\epsilon}_x) + \epsilon_y,
\end{equation*}
whose Taylor expansion centered at $x$ is
\begin{equation*}
f(\x-\boldsymbol{\epsilon}_x) = f(\x) + \boldsymbol{\epsilon}_x^\intercal \bigg(\frac{\partial f(\x)}{\partial x}\bigg) + \ldots
\end{equation*}
By doing so, we can have insight about the propagation of the input noise using the derivatives of the mean function. Doing the complete posterior distribution over this function would be computationally expensive. Instead an approximate solution can be used which closely resembles the Taylor expanded function approximation:
\begin{equation}
y = f(\x) + \boldsymbol{\epsilon}_x^\intercal\partial_{\bar{f}} + \epsilon_y,
\label{eq:error}
\end{equation}
where we denoted the vector of partial derivatives of $f$ w.r.t. the features $x^j$ as $\partial_{\bar{f}}:=[\frac{\partial f(\x)}{\partial x^1}\dots\frac{\partial f(\x)}{\partial x^D}]^\top$ and $D$ denotes the input dimension. Now we have three complete terms that can be calculated or modeled within our GP framework: (i) the predictive variance, (ii) the propagation of input noise ($\boldsymbol{\epsilon}_x^\intercal\partial_{\bar{f}}$), and (iii) the noise in the output samples $\epsilon_y$. The error in the predictions introduced by the input noise $\boldsymbol{\epsilon}_x$ can be used in addition to the predictive variance $\sigma_*^2$. Note that in this formulation the output error, $\epsilon_y$, is constant so we intentionally drop it in this work. %

The derivative of the prediction function \eqref{eq:gp_pred} in GPs only depends on the derivative of the kernel function since it is linear with respect to the $\alpha$ parameters:
\begin{align*}\label{eq:der}
	\dfrac{\partial f(\x_*)}{\partial x^j} & = 
    \dfrac{\partial {\bf k}_* \boldsymbol{\alpha}}{\partial x^j} 
    = (\partial_j {\bf k}_*)^\intercal \boldsymbol{\alpha},
\end{align*} %
where $\partial_j {\bf k}_* = [\frac{\partial K(\x_*,\x_1)}{\partial x^j},\ldots,\frac{\partial K(\x_*,\x_N)}{\partial x^j}]^\intercal$. %
In this work, we used the standard Radial Basis Function (RBF) kernel,

\begin{equation*}\label{eq:rbf_kernel}
	K(\x_n, \x_m) = \exp\left( -\frac{1}{2 \sigma^2} \|\x_n-\x_m\|_2^2 \right),
\end{equation*}
where $\sigma$ is the length-scale hyperparameter. The derivative of the RBF kernel can be easily computed analytically
\begin{equation*}\label{eq:rbf_kernel_der}
	\frac{\partial K(\x_n,\x_m)}{\partial x^j} = -\frac{(x_n^j-x_m^j)}{\sigma^2} K(\x_n, \x_m).
\end{equation*}
The standard formulation (eq. \ref{eq:gp_pred} and \ref{eq:gp_var}) takes the predictive variance without considering the input noise. However, we have shown that one could construct a formulation to consider the noise of the input samples. If one were to include this error term directly into the predictive variance, a simple formulation could be used from\cite{hutchon_gp_input_2011}.
However, this only takes into account the input noise for the training points. There are other more advanced methods which allows one to include the error term within the testing points \cite{hutchon_gp_input_2011, girard_multistep} or within the actual kernel function itself \cite{dallaire_uncertain}. The obvious drawbacks of these advanced schemes is that the complexity of the model and the computational time scales rapidly with each formulation.
In this work, we will decompose eq. \ref{eq:error} and look at the proposed input error term, $\epsilon_x^{T}\partial_{\bar{f}}$, and the predictive variance function, \ref{eq:gp_var}, to see how these terms differ in the error.

\section{Experiments}

The GPs predictive variance is a good measure of how confident the model is in the predictions given the trained data. However, this is solely related to the regions where we have more training samples\footnote{Note that \eqref{eq:gp_pred} uses the training kernel matrix as a distance metric to assess how close (similar) the test point is to the training data. The predictive variance does not exploit the training observations.}. Here we show how the input noise propagation estimation can complement the information contained in the predictive variance in order to get a better estimation of the final model error.

\subsection{Data}

We use data acquired by the IASI instrument onboard the MetOp-A satellite, which consists of 8461 spectral channels between 3.62 and 15.5 $\mu$m with a spectral sampling of 0.25 cm$^{-1}$ and a spatial resolution of 25km. We chose the 01-10-2013 for our sample space , which contained $13$ complete orbits within a $24$-hour period. Since temperature is an exemplary atmospheric parameter for weather forecasting, we used the top of the atmospheric surface temperature predictions of the European Centre for Medium-Range Weather Forecasts (ECMWF) model and the radiance IASI measurements. The IASI instrument is well characterized and the noise can be described as an additive Gaussian noise with a covariance error provided by EUMETSAT.

After the extremely noisy channels were discarded, Principal Components Analysis (PCA) was used to reduce the number of dimensions from over 4500 to 90 accounting for 99\% of the variance within data. A total of $N=5000$ points were selected as training samples from the first $5$ orbits, %
and used all 13 orbits for prediction and evaluation of the results.

\subsection{Results}

\begin{figure}[t!]
	\begin{center}
		\setlength{\tabcolsep}{-13pt}
		\hspace*{-5mm}
		\begin{tabular}{c}
			\includegraphics[width=10cm]{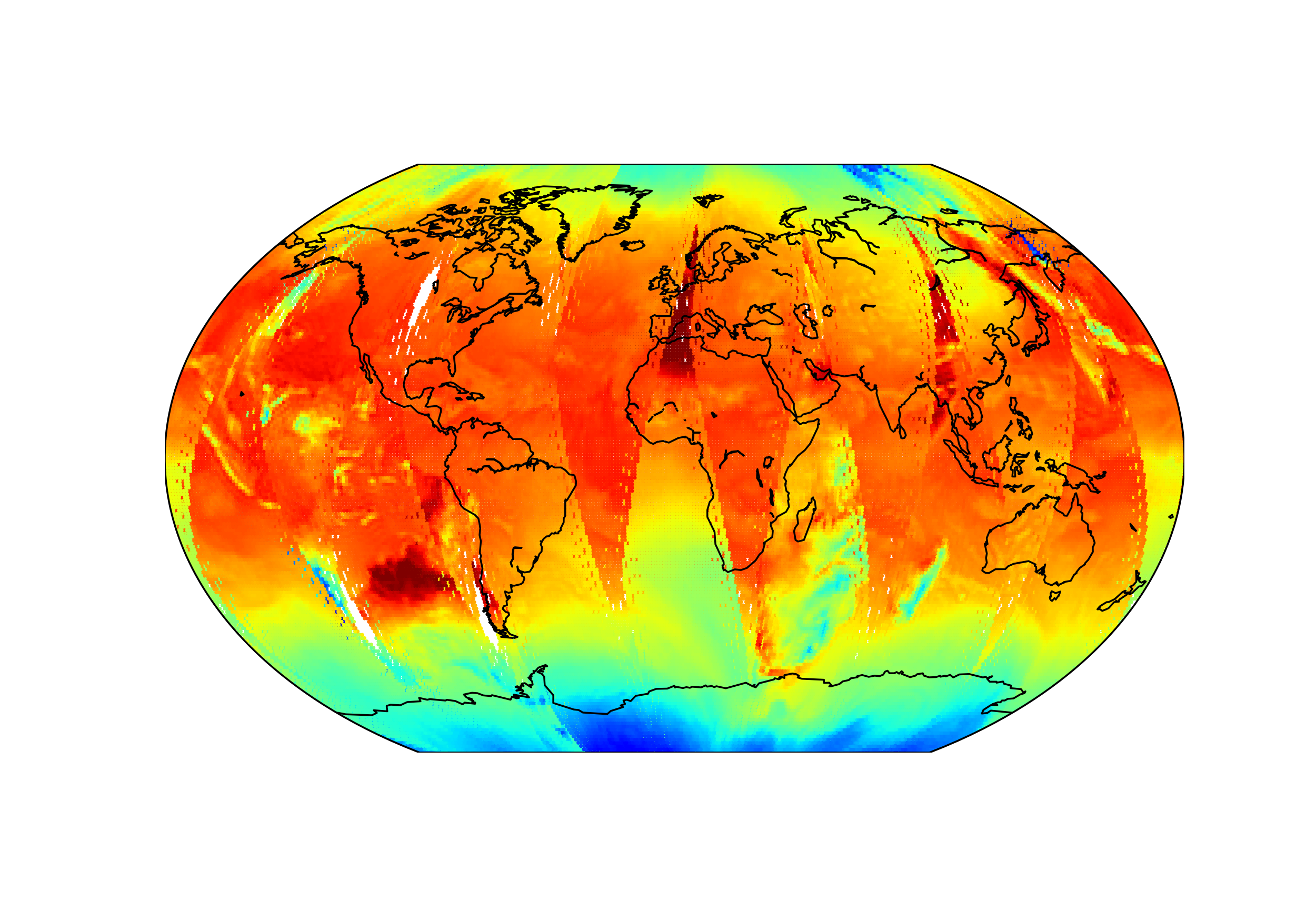} \\[-13mm]  
			a) Surface Temperature \\[-11mm]
			\includegraphics[width=10cm]{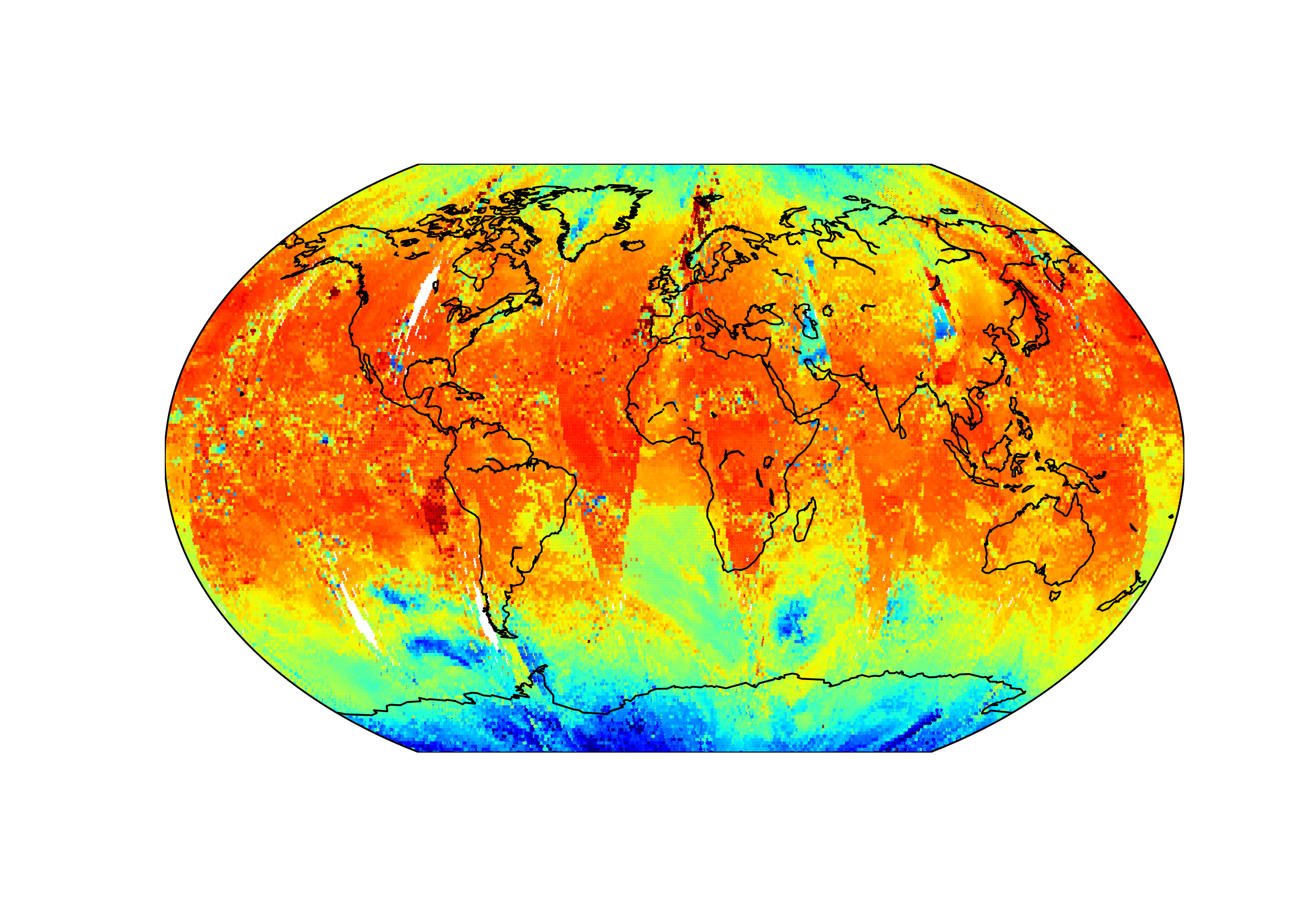} \\[-13mm]
			b) Predictions \\
            \includegraphics[width=8cm]{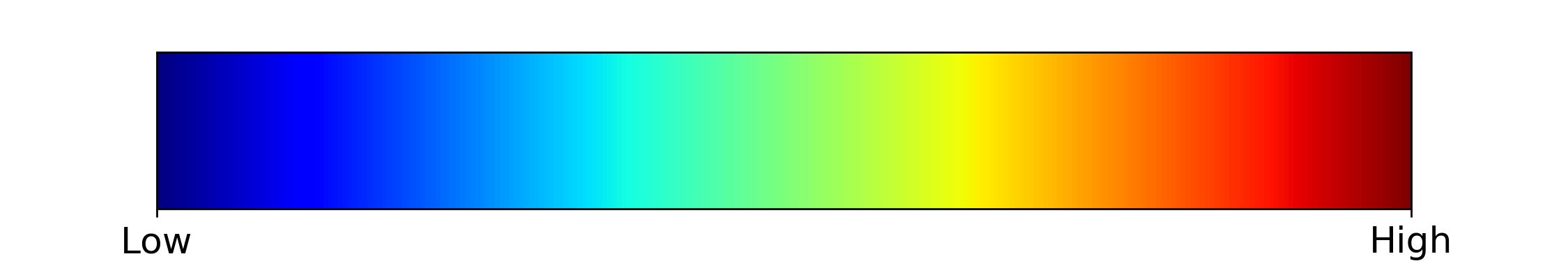}\\
		\end{tabular}
		\caption{This set of plots show the ground truth and predictions for the temperature. The blue to red color scheme represent the low to high temperatures.}
		\label{results}
	\end{center}
\end{figure}

All calculated values in this section were normalized and all final plots were transformed with histogram equalization for visualization purposes to stress the extremities for the predictive variance and error term. 
Figure \ref{results} shows the ground truth surface temperature measures provided by the ECMWF as well as the predictions of the GPs. Predictions provided by the GPs model are similar to the ground truth as they exhibit a mean absolute error value of $\ang{2.9}$C and an $R^2$ value of 0.93. However visually the model fails in some particular regions.

\begin{figure}[t!]
	\begin{center}
		\setlength{\tabcolsep}{5pt}
		\hspace*{-12mm}
		\begin{tabular}{c}
			\includegraphics[width=10cm]{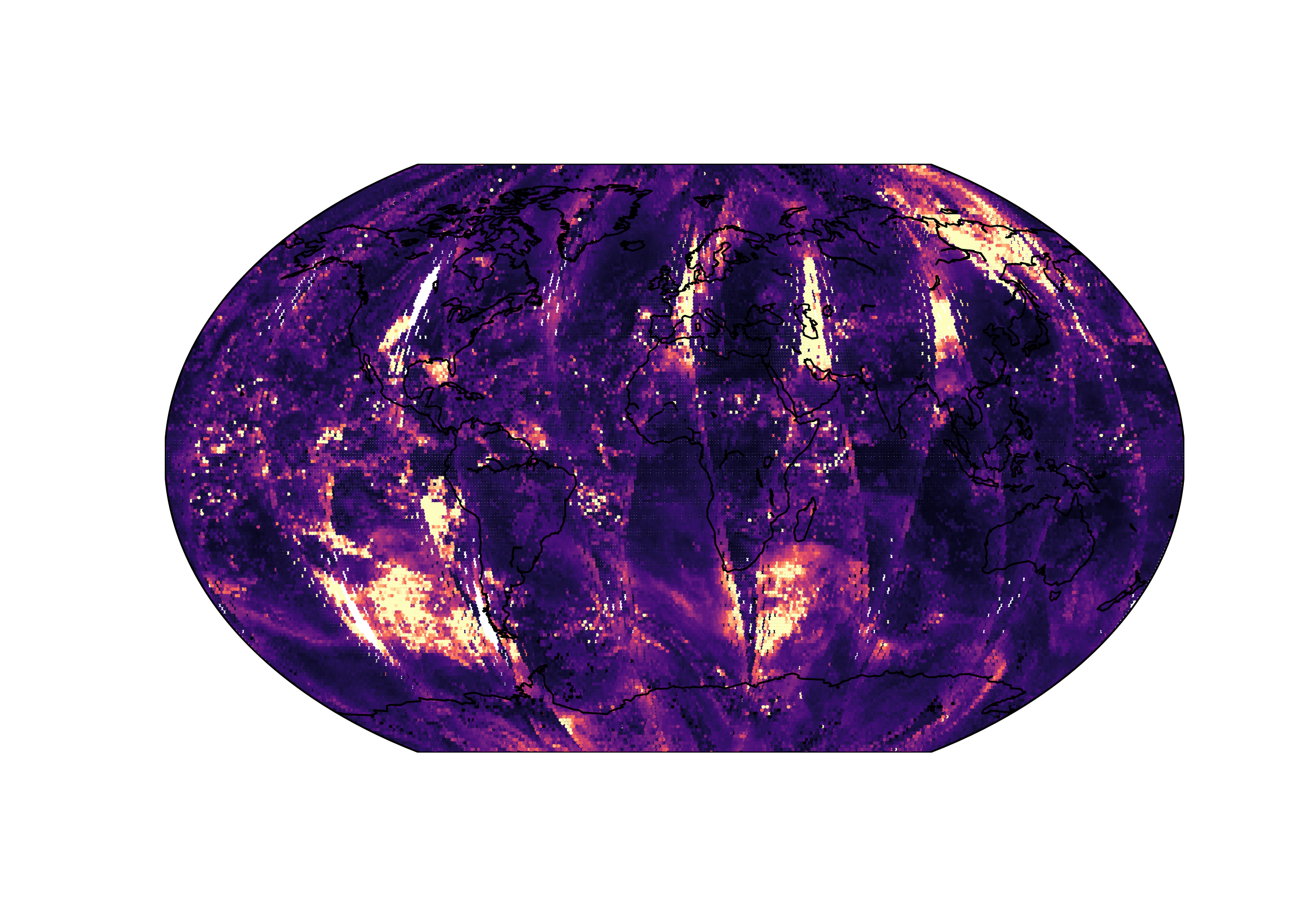} \\[-13mm]  
			a) GP Predictive Variance \\[-11mm]
			\includegraphics[width=10cm]{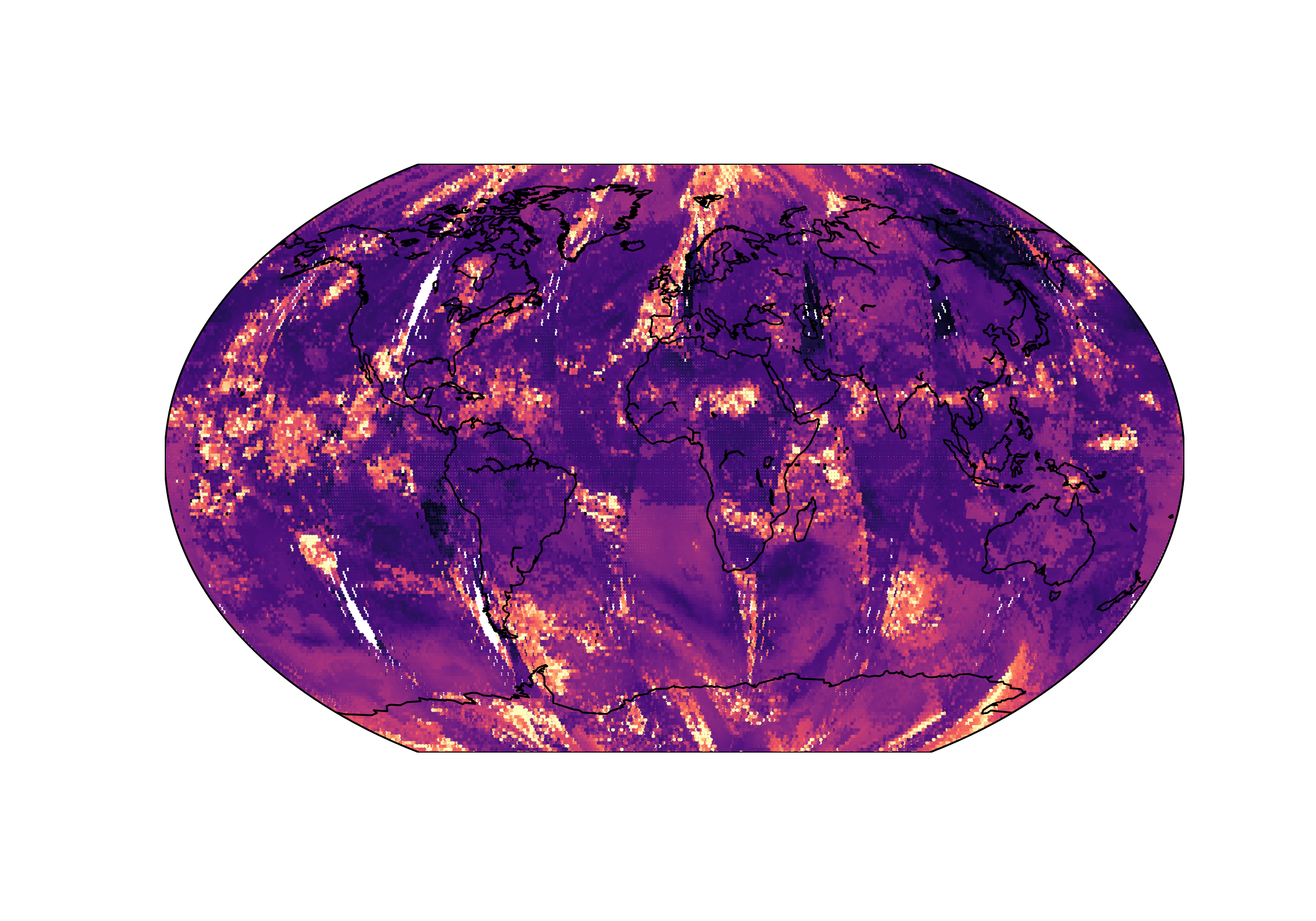} \\[-13mm]
			b) Propagated Input Noise\\[-11mm]
			\includegraphics[width=10cm]{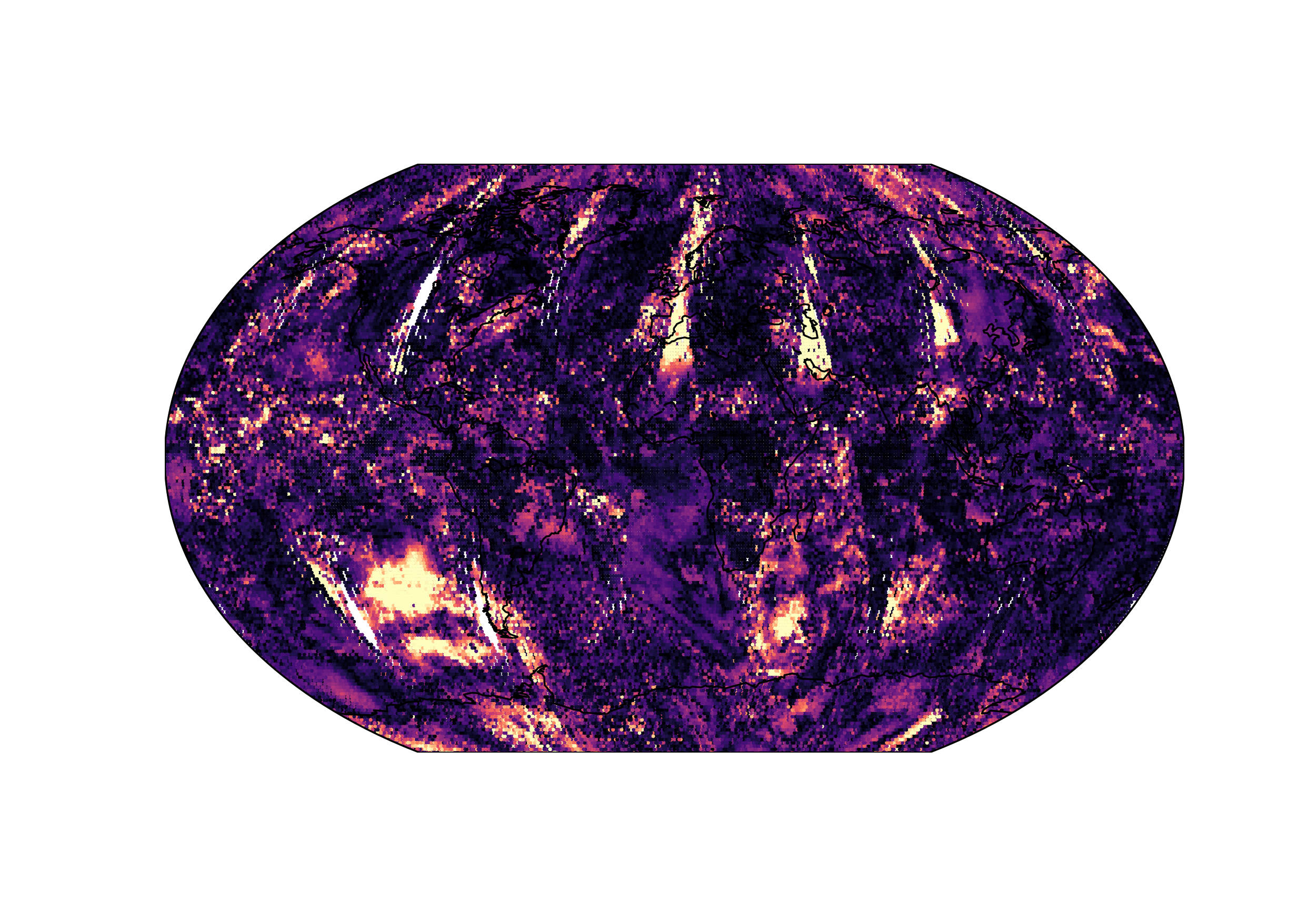}\\[-13mm]
			c) Absolute Prediction Error \\
            \includegraphics[width=10cm]{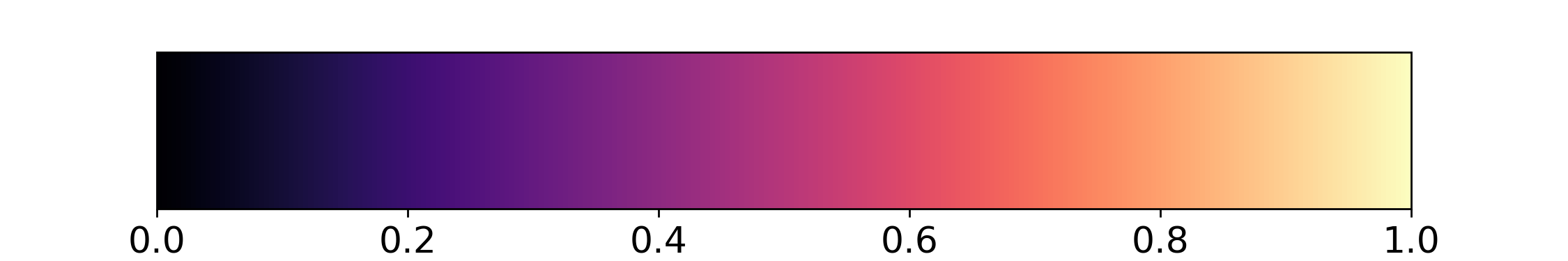}\\[2mm]
		\end{tabular}
		\caption{Comparison between a) predictive variance \eqref{eq:gp_var}, b) the propagated input noise term \eqref{eq:error},  and c) absolute prediction error for the temperature. The dark to light color scheme represent the low to high variance in temperatures for a), the low to high values of the propagated input noise term for subfigure b), and the low to high prediction errors for subfigure c).}
		\label{errors}
	\end{center}
\end{figure}

Figure \ref{errors} shows the predictive variance \eqref{eq:gp_var}, the error propagated from the learned GP model (central term in \eqref{eq:error}, as well as the absolute error of the model (i.e. the absolute value of the difference between the maps in Fig.~\ref{results}. Note that the error map is not available in a real situation since we do not have access to the actual surface temperature values. 
We can see how the information provided by the predictive variance and the propagated input noise complement each other predicting regions of the absolute prediction error map.  
The dark and light regions highlight the areas where the predictive variance is low and high respectively. %
The predictive variance corresponds to some regions within the absolute error plot; for example the south-west region of South America or the north-south streak within the Middle East. These regions do not appear on the propagated error map. However, the propagated error map showcases some regions near the poles of the globe that correspond to the absolute error but the predictive variance does not stress those regions as much. 
Furthermore, the propagated error map picks up the error in the region just over the Iberian peninsula whereas the predictive variance does not but this error does show up on the absolute error map. 

\section{Conclusion}
The consideration of input noisy is usually difficult to take into account in statistical models, however it is extremely important in Earth science communities. If we hope to combine the use of statistical models with physical models, then we will need accurate error and uncertainty estimates for our predictions. In this paper, we gave a simple formulation and rationale for how the derivative of GP models in particular can be used to help the predictive variance illustrate more accurate error estimates. 
Using a GP model to predict temperature from radiances, we showed that the predictive variance and propagated error computed using the derivative complement each other as an approximation if we compare them to the absolute error between the ground truth and the predictions. Further work would be to incorporate the input noise information both during the training procedure of the statistical methods and in the computation of the predictive variance.

\section*{Acknowledgements}

This work has been partially supported by the European Research Council (ERC) under the ERC-CoG-2014 SEDAL project (grant agreement 647423) and the Spanish Ministry of Economy and Competitiveness (MINECO, TEC2016-77741-R, ERDF

\small
\bibliographystyle{IEEEbib}
\bibliography{Zotero_eman}

\end{document}